\documentclass[letterpaper]{article} 
\usepackage[preprint]{aaai2027}  
\usepackage[hyphens]{url}  
\usepackage{graphicx} 
\usepackage{natbib}  
\usepackage{caption} 
\usepackage{algorithm}
\usepackage{algorithmic}

\usepackage{newfloat}
\usepackage{listings}
\DeclareCaptionStyle{ruled}{labelfont=normalfont,labelsep=colon,strut=off} 
\floatstyle{ruled}
\newfloat{listing}{tb}{lst}{}
\floatname{listing}{Listing}

\usepackage{booktabs}

\usepackage{amssymb}
\usepackage{amsmath}

\title{TaskSense: Focusing on What Matters in World Models}
\author {
    SM Mazharul Islam\textsuperscript{\rm 1}\corresponding,
    Manfred Huber\textsuperscript{\rm 1}
}
\affiliations {
    \textsuperscript{\rm 1}The University of Texas at Arlington\\
    smmazharul.islam@uta.edu, huber@cse.uta.com
}

\begin{document}

\maketitle

\begin{abstract}
World models for visual control typically learn compact latent states by reconstructing observations, implicitly encouraging representations to preserve information across the entire visual input. However, task-relevant content often occupies only a small fraction of the observation, while background clutter and distractors consume valuable representational capacity. This mismatch between visual reconstruction and control objectives biases latent representations to model task-irrelevant visual content, diluting learning signals for control-relevant features and severely degrading downstream performance under visual distractions. We introduce TaskSense, a task-centric world modeling framework that enforces task relevance before latent encoding through a differentiable stochastic spatial attention mechanism conditioned on the previous latent state. To steer attention toward control-relevant regions, we augment training with an auxiliary inverse-dynamics objective. Rather than reconstructing the full observation, the world model reconstructs only the attended regions, encouraging latent representations to preserve task-relevant information while discarding irrelevant visual content. The decoder is further conditioned on the sampled attention map, enabling consistent reconstruction despite stochastic attention. Compared with the DreamerV3 baseline, TaskSense maintains competitive performance on the DeepMind Control Suite while consistently outperforming DreamerV3 on the Distracting Control Suite, demonstrating substantially improved robustness to visual distractions. Qualitative analysis further confirms that the learned attention, guided by inverse-dynamics supervision, consistently localizes control-relevant regions while suppressing irrelevant visual content.
\end{abstract}


\section{Introduction}

Learning compact latent dynamics models from raw visual observations~\cite{hafner2019learning} has become a powerful paradigm for model-based reinforcement learning (MBRL)~\cite{sutton1991dyna}. By compressing high-dimensional observations into predictive latent states, world models~\cite{ha2018world} enable planning and policy optimization entirely within a learned latent space. Many successful world models use powerful decoders to reconstruct observations, making reconstruction the principal training signal for learning latent dynamics~\cite{micheli2022transformers, zhang2023storm, hafner2023mastering, hafner2025mastering}. 

Decision making, however, typically depends on only a small fraction of an observation, whereas common reconstruction objectives reward preserving all visual information. As a result, optimizing to reconstruct entire observations implicitly encourages latent representations to preserve information across the entire visual scene, regardless of whether it is relevant for control. This mismatch is often masked in standardized benchmarks~\cite{tassa2018deepmind, bellemare2013arcade}, where backgrounds are static and task-relevant objects dominate the scene. However, as observations become cluttered with distractors or task-relevant objects occupy only a small portion of the image, preserving the entire visual scene becomes increasingly inefficient, leading to substantial degradation in downstream performance~\cite{stone2021distracting, morihira2026r2}. At the opposite extreme, relying solely on task supervision such as reward prediction provides a much weaker learning signal and often fails to learn sufficiently rich visual representations~\cite{yarats2021improving}. Consequently, neither reconstructing every pixel nor learning only from rewards offers an ideal objective for representation learning.

One line of work eliminates pixel reconstruction entirely and instead learns representations through contrastive objectives~\cite{burchi2025learning}, prototype learning~\cite{deng2022dreamerpro}, or predictive consistency~\cite{morihira2026r2}. Although these methods reduce the pressure to encode irrelevant appearance, removing the decoder also sacrifices the generative world model essential for imagination-based planning.

Another line of work improves representation learning through object-centric modeling~\cite{mosbach2024sold}, feature masking~\cite{seo2023masked}, or attention mechanisms via masked reconstruction~\cite{sun2024learning}. While these methods encourage more informative latent representations, most operate after visual features have already been extracted by the encoder. Consequently, they primarily learn how to encode information rather than determining which information should enter the latent dynamics model in the first place.

\begin{figure*}[t]
\centering
\includegraphics[width=0.8\textwidth]{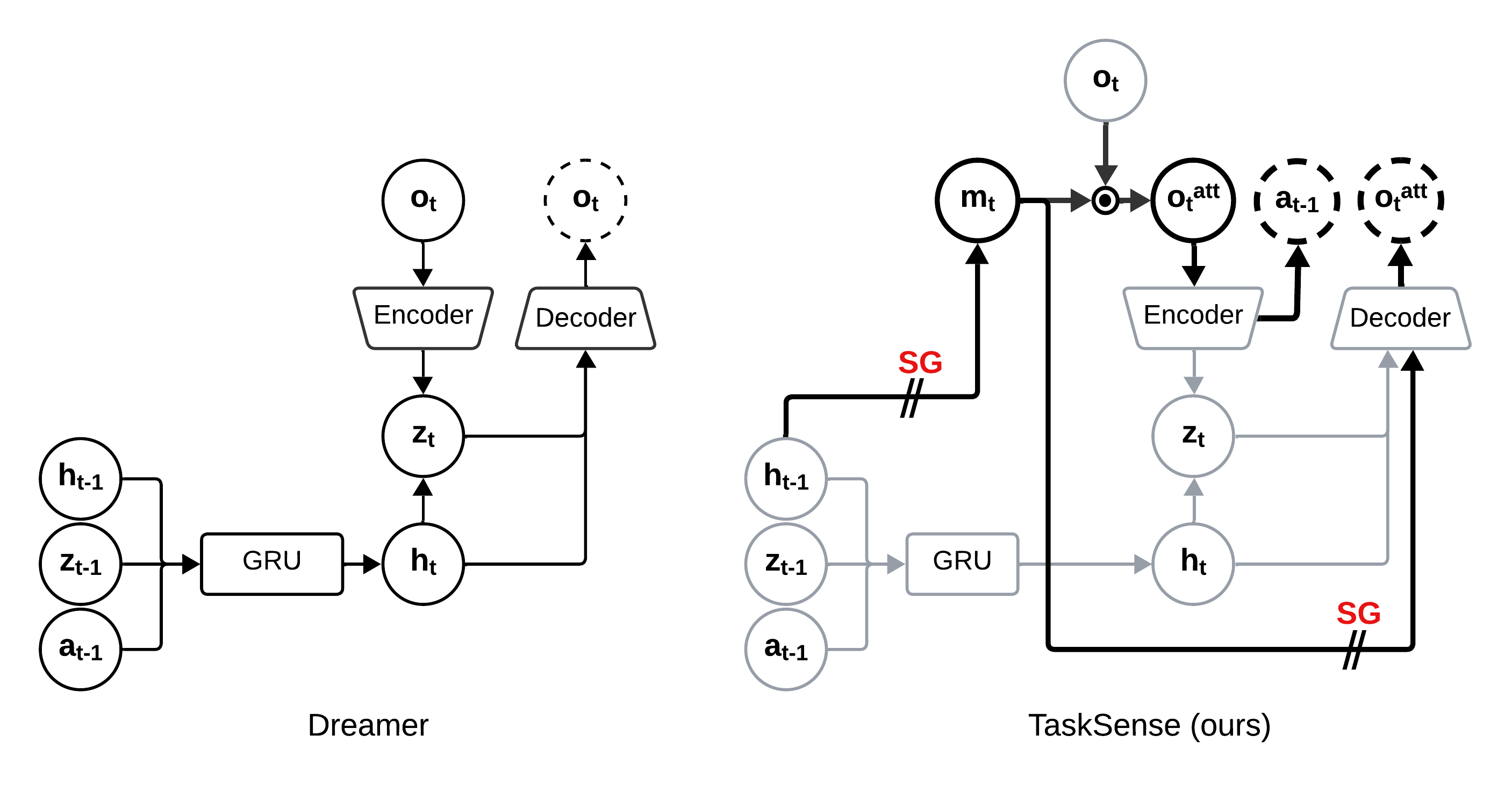} 
\caption{Comparison of Dreamer and TaskSense. Dashed circles denote predicted quantities. Left: Dreamer encodes the full observation $o_t$ into $z_t$ and reconstructs the complete observation. Right: TaskSense augments Dreamer (grayed-out components) with a stochastic attention module (black) that predicts an attention map $m_t$ from the previous belief state $h_{t-1}$ (SG). The attention map filters $o_t$ to produce $o_t^{\mathrm{att}}$, which is encoded by the original Dreamer encoder. Reconstruction is applied only to $o_t^{\mathrm{att}}$, and an auxiliary inverse-dynamics objective predicts the previous action $a_{t-1}$. SG denotes stop-gradient.}

\label{fig:overview}
\end{figure*}

A closely related direction is Task-Informed Abstraction (TIA)~\cite{fu2021learning}, which encourages task-relevant representations by jointly training two world models with cooperative reconstruction while adversarially separating one model from the reward signal. While effective, TIA explicitly learns a separate world model to capture task-irrelevant information, increasing model complexity and training overhead. This raises a fundamental question: rather than learning to separate relevant and irrelevant information after encoding, can a world model directly control which visual information enters its latent dynamics model?

We propose \textbf{TaskSense}, a task-centric world modeling framework that addresses this question by enforcing task relevance before latent encoding through latent-conditioned stochastic spatial attention. TaskSense predicts a stochastic attention map from the previous latent belief state and uses it to filter observations before encoding, allowing the model to decide where to focus based on its evolving understanding of the task rather than instantaneous visual appearance. By filtering inputs instead of modifying the latent dynamics model, TaskSense prevents task-irrelevant visual information from consuming representational capacity while preserving the reconstruction-based world model needed for imagination and planning. To avoid degenerate solutions such as suppressing all observations, TaskSense combines an auxiliary inverse-dynamics objective that encourages attention toward control-relevant regions with attention-conditioned reconstruction that preserves attended information without requiring reconstruction of irrelevant content.

Our main contributions are:

\begin{enumerate}
    \item We introduce TaskSense, a reconstruction-based world modeling framework that enforces task relevance before latent encoding through latent-conditioned stochastic spatial attention.

    \item We develop a self-supervised learning framework that grounds stochastic attention in control-relevant regions by combining inverse-dynamics supervision with reconstruction over filtered observations.

    \item We demonstrate that TaskSense matches DreamerV3 on standard DeepMind Control Suite tasks while consistently outperforming it under visual distractions. Qualitative visualizations and ablation studies further verify that inverse-dynamics supervision is essential for learning task-centric attention.
\end{enumerate}

\section{Related Work}

\subsection{World Models}

Learning latent world models has become a dominant paradigm for MBRL from high-dimensional visual observations. Early work such as World Models~\cite{ha2018world} and PlaNet~\cite{hafner2019learning} demonstrated that compact latent dynamics models enable planning directly in a learned representation space. PlaNet introduced the recurrent state-space model (RSSM), which was later adopted and extended by the Dreamer family~\cite{hafner2019dream, hafner2020mastering, hafner2023mastering, hafner2025mastering} to optimize policies and value functions through latent imagination while retaining visual reconstruction as the primary training signal. More recently, transformer-based world models such as IRIS~\cite{micheli2022transformers} and STORM~\cite{zhang2023storm} have further improved long-horizon prediction using autoregressive sequence modeling.

Despite architectural differences, these methods share a common reliance on visual reconstruction for learning latent representations, encouraging latent states to preserve information throughout the observation regardless of its relevance for control. Although highly effective in standard visual control benchmarks, this reconstruction-driven objective can also encourage latent representations to encode task-irrelevant visual content, reducing robustness under visual distractors~\cite{stone2021distracting}. TaskSense preserves the generative benefits of reconstruction-based world models while introducing task relevance before latent encoding.

\subsection{Representation Learning}

Beyond contrastive and prototype learning, several works explore alternative objectives for improving latent representations. Masked reconstruction and bisimulation objectives encourage representations that capture visual structure and behavioral equivalence, respectively~\cite{sun2024learning, zhang2020learning}. Predictive consistency and redundancy reduction improve robustness by encouraging stable latent representations while reducing reliance on data augmentations~\cite{morihira2026r2}.

Orthogonal to representation objectives, methods such as MuZero~\cite{schrittwieser2020mastering} and TD-MPC~\cite{hansen2022temporal} learn latent dynamics optimized for reward and value prediction rather than visual reconstruction.

Although these approaches improve robustness and representation quality, many reduce reliance on pixel reconstruction, sacrificing the generative observation model that enables latent imagination and planning~\cite{yarats2021improving}. In contrast, TaskSense retains reconstruction as the primary learning signal while reconstructing only task-relevant visual information.

\subsection{Learning Task-Relevant Visual Representations}

Attention mechanisms improve visual representation learning by selectively emphasizing informative regions or features. Object-centric methods explicitly decompose scenes into object-level representations, enabling reasoning over individual entities rather than holistic image features~\cite{locatello2020object, kipf2021conditional}. Recent work such as SOLD~\cite{mosbach2024sold} shows that structured object-centric representations can further improve latent world modeling and control. However, these approaches typically infer importance from visual observations or learned features rather than task objectives.

Task-driven approaches instead incorporate behavioral supervision to learn control-relevant representations. Task-Informed Abstraction (TIA)~\cite{fu2021learning} learns task-relevant visual abstractions by jointly optimizing reconstruction while separating task-irrelevant information from reward-related features. Auxiliary objectives such as inverse dynamics have also been used to encourage representations that capture controllable aspects of the environment~\cite{shelhamer2016loss, pathak2017curiosity, efroni2021provable, lamb2022guaranteed}. By predicting actions from state transitions, inverse-dynamics objectives preserve information useful for control while ignoring uncontrollable variations~\cite{agrawal2015learning}. However, these methods primarily regularize representations after visual information has been encoded rather than controlling which observations enter the latent model.

TaskSense combines task-driven supervision with selective visual processing by using inverse-dynamics supervision to ground stochastic spatial attention in control-relevant regions. Unlike prior attention mechanisms that infer importance from visual features, TaskSense predicts observation-level attention from the latent belief state and applies it before encoding, enabling reconstruction-based world models to selectively represent information relevant for control.


\section{Method}

TaskSense augments the DreamerV3~\cite{hafner2023mastering} world model with a learned stochastic spatial attention module that controls information flow before latent encoding. The attention module produces an observation mask that filters visual inputs before they are processed by the encoder, while the decoder reconstructs the filtered observations conditioned on the corresponding attention map. To prevent degenerate attention solutions, we introduce an auxiliary inverse-dynamics objective that encourages attention to preserve control-relevant visual information. Figure~\ref{fig:overview} provides an overview of the proposed architecture.

Before describing the proposed components, we briefly review the aspects of DreamerV3 relevant to our modifications. DreamerV3 employs a Recurrent State-Space Model (RSSM)~\cite{hafner2019learning}, where a GRU~\cite{cho2014properties} with block-diagonal recurrent weights~\cite{van2019rethinking} maintains a deterministic hidden state $h_t$, while stochastic latent variables are represented through a learned prior distribution $\hat{z}_t$ and posterior distribution $z_t$. During training, the posterior is inferred from the current observation $o_t$ and deterministic state $h_t$, and the decoder reconstructs observations from the latent state $s_t=(h_t,z_t)$. TaskSense retains this latent dynamics formulation and modifies only the observation interface of the world model. Specifically, TaskSense introduces a learned spatial attention mechanism that filters observations before encoding and adapts the reconstruction objective to account for stochastic attention.

\subsection{Task-Centric Spatial Attention}

Our goal is to learn a spatial attention field that filters task-irrelevant visual information before observations are encoded into the world model. Rather than predicting an attention value for every image pixel, we define attention over a coarse lattice
\[
\Lambda=\{0,\ldots,H'-1\}\times\{0,\ldots,W'-1\},
\]
covering an input observation of resolution $H\times W$. Each lattice site corresponds to an
$s\times s$ image patch, where
\[
s=\frac{H}{H'}=\frac{W}{W'}.
\]
This coarse parameterization introduces implicit local consistency by assigning a shared attention value to neighboring pixels while substantially reducing the computational overhead of dense pixel-wise attention. Thus, spatial smoothness emerges from the lattice parameterization rather than requiring additional architectural constraints.

We parameterize the attention field using a lightweight MLP $f_{\phi}$ conditioned on the previous latent belief state
$s_{t-1}=(h_{t-1},z_{t-1})$. Specifically, the attention logits and activation probabilities are computed as

\begin{equation}
    l_t=f_{\phi}\big(\mathrm{SG}(s_{t-1})\big), \qquad
    \theta_t=\sigma(l_t),
\end{equation}

where $l_t\in\mathbb{R}^{H'\times W'}$ denotes the lattice logits, $\sigma(\cdot)$ is the sigmoid function, and $\theta_{t,i,j}$ represents the probability of attending to lattice location $(i,j)$. Conditioning attention on the previous latent state allows the model to determine where to attend based on its evolving belief about the task rather than instantaneous visual appearance alone. We use $\mathrm{SG}(s_{t-1})$ instead of $s_{t-1}$ to prevent gradients from propagating through the latent dynamics state into the attention predictor, decoupling attention prediction from the optimization of the RSSM dynamics.

The resulting attention field is modeled as a collection of independent stochastic variables. Unlike normalized attention mechanisms such as softmax attention, which enforce competition among spatial locations, TaskSense allows multiple regions to be attended or ignored simultaneously. This property is important because task-relevant information may appear at multiple spatial locations and is not inherently competitive. We therefore model the attention field using a mean-field Binary Concrete (Relaxed Bernoulli) distribution,

\begin{equation}
q_{\phi}(m_t|s_{t-1})
=
\prod_{(i,j)\in\Lambda}
q_{\phi}(m_{t,i,j}|s_{t-1}).
\end{equation}

During training, differentiable attention masks are sampled using the Binary Concrete reparameterization~\cite{maddison2016concrete,jang2016categorical},

\begin{equation}
m_{t,i,j}
\sim
\mathrm{BinaryConcrete}(\theta_{t,i,j},\tau),
\label{eq:sample_attn}
\end{equation}

where $\tau$ denotes the temperature parameter. Throughout all experiments, we keep $\tau=1.0$ fixed and do not employ temperature annealing. During inference, stochastic sampling is replaced by the deterministic mean attention,
$m_{t,i,j}=\theta_{t,i,j}$.

To encourage selective attention and limit unnecessary information flow, we regularize the expected attention mass,

\begin{equation}
L_{\mathrm{sparse}}
=
\frac{1}{|\Lambda|}
\sum_{(i,j)\in\Lambda}
\theta_{t,i,j},
\label{eq:sparse_loss}
\end{equation}

where $|\Lambda|=H'W'$ is the number of lattice cells. Using the mean-field parameters instead of sampled masks avoids Monte Carlo variance in the sparsity gradients while still encouraging compact attention maps. This regularizer alone can admit degenerate solutions with excessive information removal; the auxiliary inverse-dynamics objective introduced later prevents such collapse by requiring attended observations to retain control-relevant information.

\subsection{Attention-Aware Encoding}
We upsample the sampled attention mask $m_t$ to the original observation resolution for modulating the observation through element-wise multiplication, attenuating regions assigned low attention while retaining information from attended regions.

Multiplying the observation by the attention map removes information about the attention values themselves. For example, an attenuated pixel value may correspond either to a genuinely low-intensity pixel or to a high-intensity pixel suppressed by a low attention weight. To resolve this ambiguity, we concatenate the upsampled attention map as an additional input channel,

\[
o_t^{\text{att}} = \text{Concat}\big(o_t \odot m^{\text{up}}_t, m^{\text{up}}_t\big),
\]

Note that, the same sampled attention mask is used for both observation modulation and channel concatenation. The resulting attention-aware observation is passed unchanged to the standard DreamerV3 convolutional encoder. Consequently, the only architectural modification to the encoder is the addition of a single attention channel.

\subsection{Auxiliary Inverse Dynamics}
\label{sec:aux_inv_dyn}

Since the sparsity objective encourages the attention mechanism to minimize the amount of retained visual information, a degenerate solution is to predict uniformly small attention values, simultaneously minimizing the sparsity penalty while removing meaningful signal from downstream reconstruction. Although the reinforcement learning objective provides an indirect signal for useful attention, it may be sparse and often delayed, making it insufficient to reliably prevent this collapse.

To explicitly ground attention in control-relevant information, we introduce an auxiliary inverse-dynamics objective. Intuitively, predicting the action executed between two consecutive observations encourages retaining visual features that are predictive of the agent's behavior. We therefore predict the action from the temporal difference between consecutive encoded features,

\begin{equation}
\hat{a}_t 
=
g_{\psi} \big( f_t - \mathrm{SG}(f_{t-1}) \big),
\end{equation}

where $f_t=E(o_t^{\mathrm{att}})$ denotes the encoder feature extracted from the attention-aware observation and $g_{\psi}$ is a lightweight MLP.

The previous feature is detached using a stop-gradient operator to prevent the inverse-dynamics objective from propagating gradients through the previous timestep representation. This ensures that the auxiliary objective encourages the current attention mechanism to retain action-predictive information without introducing additional dependencies across timesteps.

The inverse-dynamics loss is defined as

\begin{equation}
L_{\mathrm{inv}} = \| \hat a_t-a_t \|_2^2,
\end{equation}

which prevents collapse to trivial sparse solutions while keeping attention aligned with the downstream control objective.

\subsection{Attention-Conditioned Reconstruction}

The objective of TaskSense is to learn latent representations that preserve information relevant for control. Consequently, reconstructing the complete observation would conflict with this objective by encouraging the latent state to encode visual information intentionally removed by the attention mechanism. Instead, we reconstruct only the attention-modulated observation introduced in the previous subsection.

Because the reconstruction target depends on the sampled attention mask, the decoder must be aware of which visual regions were retained. Otherwise, an imperfect attention sample may cause the decoder to attribute missing information to the latent representation rather than to the stochastic observation filtering process. We therefore condition the decoder on the sampled attention map,

\begin{equation}
L_{\mathrm{rec}}
=
-
\log
p_\psi
\left(
o_t^{\mathrm{att}}
\mid
s_t,
\mathrm{SG}(m_t)
\right).
\end{equation}

The attention map is provided through a stop-gradient operator so that it serves as a fixed conditioning variable for reconstruction rather than an additional optimization pathway. This prevents the decoder from learning to manipulate the attention representation while allowing reconstruction to train the latent dynamics model using the information selected by attention.

\subsection{Training Objective}

TaskSense modifies the world model training objective of the DreamerV3 while leaving the policy optimization procedure unchanged. Specifically, the latent dynamics model is optimized using the attention-conditioned reconstruction objective together with the standard DreamerV3 latent dynamics losses and two additional objectives: the attention sparsity regularizer and the auxiliary inverse-dynamics loss.

The complete world model objective is defined as

\begin{equation}
\mathcal{L}_{\mathrm{WM}}
=
\mathcal{L}_{\mathrm{dyn}}
+
\mathcal{L}_{\mathrm{rec}}
+
\lambda_{\mathrm{sparse}}\mathcal{L}_{\mathrm{sparse}}
+
\lambda_{\mathrm{inv}}\mathcal{L}_{\mathrm{inv}},
\end{equation}

where $\mathcal{L}_{\mathrm{dyn}}$ denotes the standard DreamerV3 latent dynamics objective, including the RSSM transition and representation losses. $\mathcal{L}_{\mathrm{rec}}$ corresponds to the attention-conditioned reconstruction objective described previously, while $\mathcal{L}_{\mathrm{sparse}}$ and $\mathcal{L}_{\mathrm{inv}}$ encourage compact attention maps and preservation of action-predictive visual information, respectively. We set $\lambda_{\mathrm{sparse}}=0.1$ and $\lambda_{\mathrm{inv}}=1.0$ in all experiments.

The actor-critic optimization remains unchanged and is trained using imagined trajectories generated from the learned latent dynamics model following the standard DreamerV3 procedure.

\section{Experiments}

We evaluate TaskSense on standard and distraction-rich visual control benchmarks to assess its effectiveness, analyze the learned attention mechanism, and quantify the contribution of selected design component. Specifically, we seek to answer the following questions:

\begin{itemize}
    \item[(Q1)] Does TaskSense improve robustness to visual distractions while maintaining performance in standard visual control?
    \item[(Q2)] Does the learned stochastic attention consistently identify control-relevant visual regions?
    \item[(Q3)] How important are the individual components of TaskSense?
\end{itemize}

\begin{table}[t]
\centering
\small
\begin{tabular}{lcc}
  \toprule
  Name & Symbol & Value \\
  \midrule
  Attention lattice size & $(H',W')$ & $8\times8$ \\
  Binary Concrete temperature & $\tau$ & $1.0$ \\
  Sparsity coefficient & $\lambda_{\mathrm{sparse}}$ & $0.1$ \\
  Inverse dynamics coefficient & $\lambda_{\mathrm{inv}}$ & $1.0$ \\
  \bottomrule
\end{tabular}
\caption{TaskSense-specific hyperparameters. All remaining hyperparameters follow the official DreamerV3 implementation.}
\label{table:hyperparam}
\end{table}

\begin{table*}[t]
\centering
\begin{tabular}{lcccc}
  \toprule
  & \multicolumn{2}{c}{DeepMind Control Suite} & \multicolumn{2}{c}{Distracting Control Suite} \\
  \cmidrule(lr){2-3} \cmidrule(lr){4-5}
  Task & TaskSense & DreamerV3 & TaskSense & DreamerV3 \\
  \midrule
  Cartpole Swingup  & $\mathbf{865.0}$ & $849.2$ & $\mathbf{140.3}$ & $124.6$ \\
  Cheetah Run    & $754.9$ & $\mathbf{885.0}$ & $\mathbf{401.6}$ & $267.3$ \\
  Hopper Hop   & $220.9$ & $\mathbf{265.8}$ & $\mathbf{1.9}$ & $0.7$ \\
  Walker Stand & $\mathbf{977.1}$ & $976.8$ & $\mathbf{972.2}$ & $965.6$ \\
  Walker Walk   & $958.8$ & $\mathbf{963.0}$ & $\mathbf{893.8}$ & $737.8$ \\
  Walker Run   & $474.0$ & $\mathbf{723.9}$ & $\mathbf{531.2}$ & $286.2$ \\
  \midrule
  Task Mean  & $708.4$ & $\mathbf{777.3}$ & $\mathbf{490.2}$ & $397.03$ \\
  Task Median & $810.0$ & $\mathbf{867.1}$ & $\mathbf{466.4}$ & $276.8$ \\
  \bottomrule
\end{tabular}
\caption{Average episode returns. TaskSense matches DreamerV3 on standard DeepMind Control Suite tasks while substantially improving robustness on the distracting benchmark with dynamic natural backgrounds.}
\label{table:comparison}
\end{table*}

\subsection{Experimental Setup}

\subsubsection{Environments}

We evaluate TaskSense on the DeepMind Control Suite (DMC)~\cite{tassa2018deepmind} and the Distracting Control Suite (DCS)~\cite{stone2021distracting}. All experiments use RGB observations of resolution $64\times64$.

The DeepMind Control Suite provides standard continuous-control benchmarks with static backgrounds. The Distracting Control Suite augments the same tasks with dynamic natural-image backgrounds from the DAVIS dataset~\cite{Pont-Tuset_arXiv_2017}, providing a challenging benchmark for evaluating robustness to visual distractions. Unlike the original benchmark, which plays DAVIS videos sequentially, we independently sample a background frame at every environment step while keeping the camera viewpoint and agent colors unchanged. This removes temporal consistency in the background and produces a substantially more challenging distraction setting.

Experiments are conducted on six representative DMC tasks: \textit{Cartpole Swingup}, \textit{Cheetah Run}, \textit{Hopper Hop}, \textit{Walker Stand}, \textit{Walker Walk}, and \textit{Walker Run}. These tasks span diverse dynamics and varying levels of control difficulty, covering both locomotion and manipulation behaviors.

\subsubsection{Implementation Details}

We build TaskSense by extending the official DreamerV3 implementation. Unless otherwise stated, TaskSense and DreamerV3 share identical network architectures, optimization procedures, replay buffers, and training schedules. All hyperparameters not specific to TaskSense follow the official DreamerV3 implementation. The only additions are the observation-level stochastic attention module, attention-aware reconstruction objective, and auxiliary inverse-dynamics objective described in the method section. The attention predictor and inverse-dynamics head are implemented as 3-layer MLPs with 256 units per hidden layer. Table~\ref{table:hyperparam} summarizes the additional TaskSense-specific hyperparameters.

\subsubsection{Training and Evaluation}

All experiments are repeated with five random seeds. Each agent is trained for $500$k environment steps, and evaluation is performed every $10$k training steps. At each evaluation point, the agent is evaluated over 10 episodes, with each episode having a maximum length of 1,000 environment steps. The reported performance is the average episode return across evaluation episodes and random seeds. Following previous DreamerV3 studies~\cite{hafner2023mastering}, we use an action repeat of 2 for all environments. All experiments are conducted on a single NVIDIA RTX A6000 GPU, with each $500$k environment-step training run requiring approximately 4.5 hours.

\begin{figure}[t]
\centering
\includegraphics[width=0.45\textwidth]{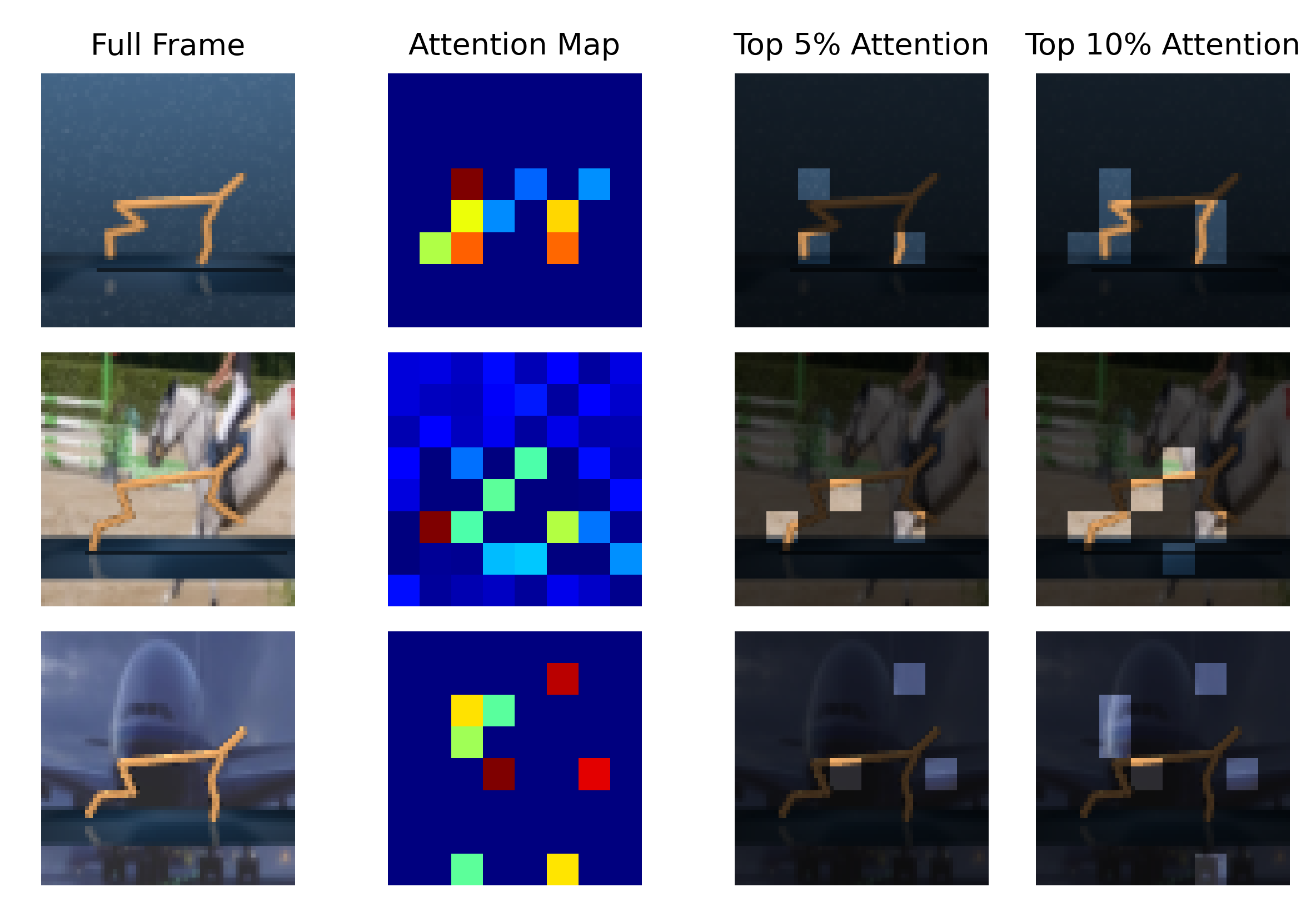} 
\caption{Visualization of the learned stochastic spatial attention on the \textit{Cheetah Run} task. The first row shows the standard DeepMind Control Suite, the second row shows the corresponding distracting environment with dynamic natural backgrounds, and the third row shows TaskSense without the auxiliary inverse-dynamics objective. From left to right, we show the original observation, predicted attention map, and reconstructions using the top $5\%$ and top $10\%$ attended regions. TaskSense consistently focuses on the cheetah's legs while suppressing irrelevant visual content, whereas removing inverse-dynamics supervision leads to diffuse and less localized attention.}
\label{fig:attention_vis}
\end{figure}

\begin{figure*}[t]
\centering
\includegraphics[width=0.9\textwidth]{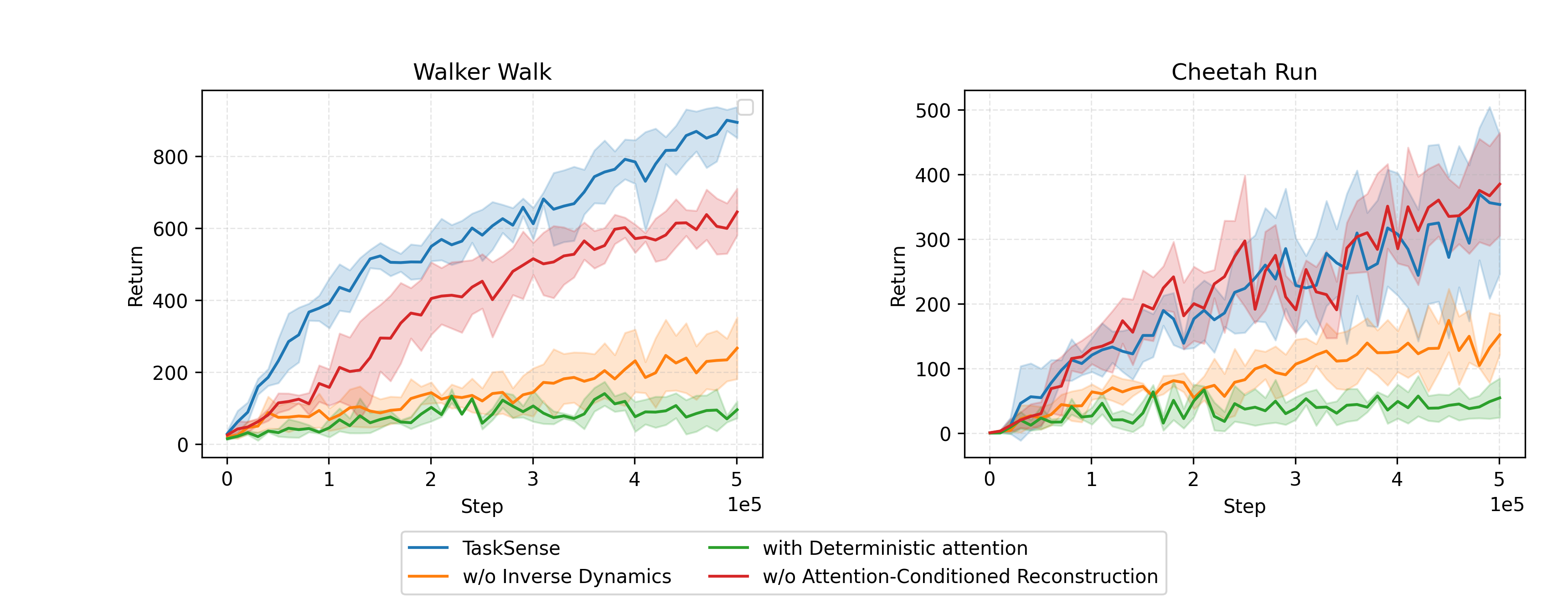} 
\caption{Ablation study on the Distracting Control Suite for \textit{Walker Walk} and \textit{Cheetah Run}. We compare TaskSense with three variants: removing the inverse-dynamics objective, replacing stochastic attention with deterministic attention, and removing attention-conditioned reconstruction. Each curve shows the mean return over five random seeds, with shaded regions indicating one standard deviation.}
\label{fig:ablation}
\end{figure*}

\subsection{Performance on Visual Control Benchmarks}

We first evaluate whether TaskSense improves robustness to visual distractions while maintaining performance in standard visual control. We compare TaskSense with DreamerV3 on the DeepMind Control Suite (DMC) and Distracting Control Suite under identical training budgets. These benchmarks assess complementary properties: DMC evaluates whether observation filtering removes task-relevant information, while the distracting benchmark measures robustness when observations contain substantial irrelevant visual content.

As shown in Table~\ref{table:comparison}, TaskSense matches DreamerV3 on nearly all standard DMC tasks despite reconstructing only attention-filtered observations rather than full images. This demonstrates that selectively encoding task-relevant information is sufficient for effective latent planning without preserving every visual detail. Under distracting natural backgrounds, TaskSense consistently outperforms DreamerV3 across all evaluated tasks. By filtering irrelevant visual content before it enters the latent dynamics model, TaskSense reduces the influence of nuisance information on learned representations and improves control performance.

These results support our hypothesis that reconstruction-based world models do not need to preserve every pixel of an observation. Instead, enforcing task relevance before latent encoding enables efficient allocation of representational capacity toward control-relevant information while remaining compatible with reconstruction-based imagination.

\subsection{Qualitative Analysis}

Figure~\ref{fig:attention_vis} visualizes TaskSense's stochastic spatial attention on the \textit{Cheetah Run} task across standard DMC, distracting environments with dynamic natural backgrounds, and a model without the inverse-dynamics objective. We show the predicted attention maps and the image regions corresponding to the top $5\%$ and $10\%$ attention values. TaskSense consistently focuses on the cheetah's legs while suppressing the torso, static scene elements, and distracting backgrounds. This behavior across both standard and distracting settings indicates that attention is guided by control relevance rather than visual saliency. The inverse-dynamics objective is essential for this behavior: while sparsity encourages compact attention, it does not indicate which information is useful for control. Without inverse dynamics, attention becomes diffuse and less aligned with locomotion dynamics.

These results demonstrate that task-conditioned observation filtering enables the world model to preserve control-relevant visual information while suppressing irrelevant variations, providing an effective interface for reconstruction-based world models.

\subsection{Ablation Study}

We evaluate the contribution of TaskSense components on distracting \textit{Walker Walk} and \textit{Cheetah Run}. We compare the full model against three variants: (1) removing inverse-dynamics supervision, (2) replacing stochastic attention with deterministic attention, and (3) removing attention-conditioned reconstruction by reconstructing full observations. Figure~\ref{fig:ablation} reports the corresponding learning curves.

Replacing stochastic attention with deterministic attention causes a significant performance drop on both tasks. Stochastic attention exposes the model to diverse partial observations during training, producing representations that remain robust under varying visual conditions.

Removing the inverse-dynamics objective also substantially degrades performance, as attention lacks explicit task-related supervision and becomes less aligned with control-relevant regions. This is consistent with Figure~\ref{fig:attention_vis}, where attention without inverse dynamics fails to consistently focus on the agent's limbs.

Removing attention-conditioned reconstruction leads to a smaller but noticeable degradation, particularly on \textit{Walker Walk}. This suggests that aligning reconstruction with attended observations helps prevent the reconstruction objective from encouraging irrelevant visual information.

Overall, the ablations show that TaskSense improvements arise from the combination of stochastic attention, task-specific supervision, and attention-aligned reconstruction rather than observation masking alone.

\section{Limitations}
Throughout this work, we treat task-relevant and control-relevant visual information as interchangeable, assuming that visual features predictive of the agent's actions are also those most relevant for solving the downstream control task. Under this assumption, inverse dynamics provides effective supervision for learning task-centric attention. However, when action-predictive cues differ from long-term reward-relevant information, inverse dynamics may focus attention on suboptimal features. Developing supervisory signals that better capture long-horizon task relevance remains an important direction for future work.


\section{Conclusion}
We introduced TaskSense, a task-centric world modeling framework that enforces task relevance before latent encoding through stochastic spatial attention. By filtering observations prior to encoding and reconstructing only attended regions, TaskSense reduces representational capacity spent on task-irrelevant visual content while preserving the benefits of reconstruction-based training.

Experimental results show that TaskSense matches DreamerV3 on standard DeepMind Control Suite tasks while consistently improving robustness under visual distractions. Qualitative visualizations demonstrate that the learned attention focuses on control-relevant regions while suppressing distracting backgrounds. Ablation studies further show that both stochastic attention and inverse-dynamics supervision are essential for learning effective task-centric attention. Together, these findings suggest that enforcing task relevance at the observation interface is a promising direction for improving visual world models and representation learning in model-based reinforcement learning.


\bibliography{aaai2027}

@article{jang2016categorical,
  title={Categorical reparameterization with gumbel-softmax},
  author={Jang, Eric and Gu, Shixiang and Poole, Ben},
  journal={arXiv preprint arXiv:1611.01144},
  year={2016}
}

@article{hafner2019dream,
  title={Dream to control: Learning behaviors by latent imagination},
  author={Hafner, Danijar and Lillicrap, Timothy and Ba, Jimmy and Norouzi, Mohammad},
  journal={arXiv preprint arXiv:1912.01603},
  year={2019}
}

@article{hansen2022temporal,
  title={Temporal difference learning for model predictive control},
  author={Hansen, Nicklas and Wang, Xiaolong and Su, Hao},
  journal={arXiv preprint arXiv:2203.04955},
  year={2022}
}

@article{schrittwieser2020mastering,
  title={Mastering atari, go, chess and shogi by planning with a learned model},
  author={Schrittwieser, Julian and Antonoglou, Ioannis and Hubert, Thomas and Simonyan, Karen and Sifre, Laurent and Schmitt, Simon and Guez, Arthur and Lockhart, Edward and Hassabis, Demis and Graepel, Thore and others},
  journal={Nature},
  volume={588},
  number={7839},
  pages={604--609},
  year={2020},
  publisher={Nature Publishing Group UK London}
}

@article{zhang2020learning,
  title={Learning invariant representations for reinforcement learning without reconstruction},
  author={Zhang, Amy and McAllister, Rowan and Calandra, Roberto and Gal, Yarin and Levine, Sergey},
  journal={arXiv preprint arXiv:2006.10742},
  year={2020}
}

@article{hafner2020mastering,
  title={Mastering atari with discrete world models},
  author={Hafner, Danijar and Lillicrap, Timothy and Norouzi, Mohammad and Ba, Jimmy},
  journal={arXiv preprint arXiv:2010.02193},
  year={2020}
}

@article{micheli2022transformers,
  title={Transformers are sample-efficient world models},
  author={Micheli, Vincent and Alonso, Eloi and Fleuret, Fran{\c{c}}ois},
  journal={arXiv preprint arXiv:2209.00588},
  year={2022}
}

@article{zhang2023storm,
  title={Storm: Efficient stochastic transformer based world models for reinforcement learning},
  author={Zhang, Weipu and Wang, Gang and Sun, Jian and Yuan, Yetian and Huang, Gao},
  journal={Advances in Neural Information Processing Systems},
  volume={36},
  pages={27147--27166},
  year={2023}
}

@inproceedings{cho2014properties,
  title={On the properties of neural machine translation: Encoder--decoder approaches},
  author={Cho, Kyunghyun and Van Merri{\"e}nboer, Bart and Bahdanau, Dzmitry and Bengio, Yoshua},
  booktitle={Proceedings of SSST-8, eighth workshop on syntax, semantics and structure in statistical translation},
  pages={103--111},
  year={2014}
}

@inproceedings{yarats2021improving,
  title={Improving sample efficiency in model-free reinforcement learning from images},
  author={Yarats, Denis and Zhang, Amy and Kostrikov, Ilya and Amos, Brandon and Pineau, Joelle and Fergus, Rob},
  booktitle={Proceedings of the aaai conference on artificial intelligence},
  volume={35},
  number={12},
  pages={10674--10681},
  year={2021}
}

@article{Pont-Tuset_arXiv_2017,
  author = {Jordi Pont-Tuset and Federico Perazzi and Sergi Caelles and Pablo Arbel\'aez and Alexander Sorkine-Hornung and Luc {Van Gool}},
  title = {The 2017 DAVIS Challenge on Video Object Segmentation},
  journal = {arXiv:1704.00675},
  year = {2017}
}

@inproceedings{agrawal2015learning,
  title={Learning to see by moving},
  author={Agrawal, Pulkit and Carreira, Joao and Malik, Jitendra},
  booktitle={Proceedings of the IEEE international conference on computer vision},
  pages={37--45},
  year={2015}
}

@inproceedings{fu2021learning,
  title={Learning task informed abstractions},
  author={Fu, Xiang and Yang, Ge and Agrawal, Pulkit and Jaakkola, Tommi},
  booktitle={International Conference on Machine Learning},
  pages={3480--3491},
  year={2021},
  organization={PMLR}
}

@article{van2019rethinking,
  title={Rethinking full connectivity in recurrent neural networks},
  author={Van Keirsbilck, Matthijs and Keller, Alexander and Yang, Xiaodong},
  journal={arXiv preprint arXiv:1905.12340},
  year={2019}
}

@article{efroni2021provable,
  title={Provable rl with exogenous distractors via multistep inverse dynamics},
  author={Efroni, Yonathan and Misra, Dipendra and Krishnamurthy, Akshay and Agarwal, Alekh and Langford, John},
  journal={arXiv preprint arXiv:2110.08847},
  year={2021}
}

@article{lamb2022guaranteed,
  title={Guaranteed discovery of control-endogenous latent states with multi-step inverse models},
  author={Lamb, Alex and Islam, Riashat and Efroni, Yonathan and Didolkar, Aniket and Misra, Dipendra and Foster, Dylan and Molu, Lekan and Chari, Rajan and Krishnamurthy, Akshay and Langford, John},
  journal={arXiv preprint arXiv:2207.08229},
  year={2022}
}

@inproceedings{kipf2021conditional,
  title={Conditional object-centric learning from video},
  author={Kipf, Thomas and Elsayed, Gamaleldin Fathy and Mahendran, Aravindh and Stone, Austin and Sabour, Sara and Heigold, Georg and Jonschkowski, Rico and Dosovitskiy, Alexey and Greff, Klaus},
  booktitle={International conference on learning representations},
  year={2021}
}

@inproceedings{seo2023masked,
  title={Masked world models for visual control},
  author={Seo, Younggyo and Hafner, Danijar and Liu, Hao and Liu, Fangchen and James, Stephen and Lee, Kimin and Abbeel, Pieter},
  booktitle={Conference on Robot Learning},
  pages={1332--1344},
  year={2023},
  organization={PMLR}
}

@article{shelhamer2016loss,
  title={Loss is its own reward: Self-supervision for reinforcement learning},
  author={Shelhamer, Evan and Mahmoudieh, Parsa and Argus, Max and Darrell, Trevor},
  journal={arXiv preprint arXiv:1612.07307},
  year={2016}
}

@inproceedings{pathak2017curiosity,
  title={Curiosity-driven exploration by self-supervised prediction},
  author={Pathak, Deepak and Agrawal, Pulkit and Efros, Alexei A and Darrell, Trevor},
  booktitle={International conference on machine learning},
  pages={2778--2787},
  year={2017},
  organization={PMLR}
}

@article{locatello2020object,
  title={Object-centric learning with slot attention},
  author={Locatello, Francesco and Weissenborn, Dirk and Unterthiner, Thomas and Mahendran, Aravindh and Heigold, Georg and Uszkoreit, Jakob and Dosovitskiy, Alexey and Kipf, Thomas},
  journal={Advances in neural information processing systems},
  volume={33},
  pages={11525--11538},
  year={2020}
}

@article{sun2024learning,
  title={Learning latent dynamic robust representations for world models},
  author={Sun, Ruixiang and Zang, Hongyu and Li, Xin and Islam, Riashat},
  journal={arXiv preprint arXiv:2405.06263},
  year={2024}
}

@article{mosbach2024sold,
  title={Sold: Slot object-centric latent dynamics models for relational manipulation learning from pixels},
  author={Mosbach, Malte and Ewertz, Jan Niklas and Villar-Corrales, Angel and Behnke, Sven},
  journal={arXiv preprint arXiv:2410.08822},
  year={2024}
}

@article{sutton1991dyna,
  title={Dyna, an integrated architecture for learning, planning, and reacting},
  author={Sutton, Richard S},
  journal={ACM Sigart Bulletin},
  volume={2},
  number={4},
  pages={160--163},
  year={1991},
  publisher={ACM New York, NY, USA}
}

@article{ha2018world,
  title={World models},
  author={Ha, David and Schmidhuber, J{\"u}rgen},
  journal={arXiv preprint arXiv:1803.10122},
  year={2018}
}

@article{hafner2023mastering,
  title={Mastering diverse domains through world models, 2023},
  author={Hafner, Danijar and Pasukonis, Jurgis and Ba, Jimmy and Lillicrap, Timothy},
  journal={URL https://arxiv. org/abs/2301.04104},
  year={2023}
}

@inproceedings{hafner2019learning,
  title={Learning latent dynamics for planning from pixels},
  author={Hafner, Danijar and Lillicrap, Timothy and Fischer, Ian and Villegas, Ruben and Ha, David and Lee, Honglak and Davidson, James},
  booktitle={International conference on machine learning},
  pages={2555--2565},
  year={2019},
  organization={PMLR}
}

@article{burchi2025learning,
  title={Learning transformer-based world models with contrastive predictive coding},
  author={Burchi, Maxime and Timofte, Radu},
  journal={arXiv preprint arXiv:2503.04416},
  year={2025}
}

@article{hafner2025mastering,
  title={Mastering diverse control tasks through world models},
  author={Hafner, Danijar and Pasukonis, Jurgis and Ba, Jimmy and Lillicrap, Timothy},
  journal={Nature},
  volume={640},
  number={8059},
  pages={647--653},
  year={2025},
  publisher={Nature Publishing Group UK London}
}

@article{maddison2016concrete,
  title={The concrete distribution: A continuous relaxation of discrete random variables},
  author={Maddison, Chris J and Mnih, Andriy and Teh, Yee Whye},
  journal={arXiv preprint arXiv:1611.00712},
  year={2016}
}

@article{tassa2018deepmind,
  title={Deepmind control suite},
  author={Tassa, Yuval and Doron, Yotam and Muldal, Alistair and Erez, Tom and Li, Yazhe and Casas, Diego de Las and Budden, David and Abdolmaleki, Abbas and Merel, Josh and Lefrancq, Andrew and others},
  journal={arXiv preprint arXiv:1801.00690},
  year={2018}
}

@article{bellemare2013arcade,
  title={The arcade learning environment: An evaluation platform for general agents},
  author={Bellemare, Marc G and Naddaf, Yavar and Veness, Joel and Bowling, Michael},
  journal={Journal of artificial intelligence research},
  volume={47},
  pages={253--279},
  year={2013}
}

@article{stone2021distracting,
  title={The Distracting Control Suite--A Challenging Benchmark for Reinforcement Learning from Pixels},
  author={Stone, Austin and Ramirez, Oscar and Konolige, Kurt and Jonschkowski, Rico},
  journal={arXiv preprint arXiv:2101.02722},
  year={2021}
}

@article{morihira2026r2,
  title={R2-Dreamer: Redundancy-reduced world models without decoders or augmentation},
  author={Morihira, Naoki and Nahar, Amal and Bharadwaj, Kartik and Kato, Yasuhiro and Hayashi, Akinobu and Harada, Tatsuya},
  journal={arXiv preprint arXiv:2603.18202},
  year={2026}
}

@inproceedings{deng2022dreamerpro,
  title={Dreamerpro: Reconstruction-free model-based reinforcement learning with prototypical representations},
  author={Deng, Fei and Jang, Ingook and Ahn, Sungjin},
  booktitle={International conference on machine learning},
  pages={4956--4975},
  year={2022},
  organization={PMLR}
}


\end{document}